\documentclass{article}

\usepackage{spconf,amsmath,graphicx}
\usepackage{hyperref}
\usepackage{multirow}
\usepackage{cite}
\usepackage{array}
\usepackage{xcolor}
\usepackage{booktabs}
\usepackage{makecell}
\usepackage{siunitx}
\usepackage{psfrag}

\title{On the role of visual cues in audiovisual speech enhancement}

\name{\begin{tabular}{c}Zakaria Aldeneh, Anushree Prasanna Kumar, Barry-John Theobald, Erik Marchi,\\ Sachin Kajarekar, Devang Naik, Ahmed Hussen Abdelaziz\end{tabular}}
\address{Apple, Cupertino, CA, USA}

\begin{document}

\maketitle
%

\begin{abstract}
We present an introspection of an audiovisual speech enhancement model. 
In particular, we focus on interpreting how a neural audiovisual speech enhancement model uses visual cues to improve the quality of the target speech signal.
We show that visual cues provide not only high-level information about speech activity, i.e.,\ speech/silence, but also fine-grained visual information about the place of articulation. 
One byproduct of this finding is that the learned visual embeddings can be used as features for other visual speech applications.
We demonstrate the effectiveness of the learned visual embeddings for classifying visemes (the visual analogy to phonemes).
Our results provide insight into important aspects of audiovisual speech enhancement and demonstrate how such models can be used for self-supervision tasks for visual speech applications.
\end{abstract}
\begin{keywords}
audiovisual speech enhancement, \\lip reading, viseme classification, self-supervised learning
\end{keywords}

\section{Introduction}\label{sec_intro}

The goal of monaural (single-channel) speech enhancement is to improve the quality and intelligibility of speech when the audio is recorded in a noisy environment from a single microphone~\cite{wang2018supervised}. Enhancement models attenuate additive noise from a speech signal and can be used as pre-processors for various downstream applications, including automatic speech recognition (ASR) and speaker verification~\cite{sivasankaran2015robust,moore2017speech,shon2019voiceid,tan2018convolutional,mamun2019convolutional}.   

Previous research has shown that acoustic models used for speech enhancement benefit from the addition of visual cues~\cite{Abdelaziz13,Zeiler2016IntroducingTT,gabbay2017visual,afouras2018conversation,ephrat2018looking,afouras2019my}. Although these models have shown promising results, it is unclear how visual cues are utilized by the models. One hypothesis is that visual cues only provide high-level information about speech activity, i.e.,\ speech vs.~no speech, depending on whether the lips are moving or not.  An alternative hypothesis is that visual cues provide fine-grained information about what is being articulated.  Our work aims to interpret how visual cues are used by audiovisual speech enhancement models.  Such analysis is not only necessary for understanding the mechanism by which an audiovisual enhancer uses visual cues, but also for understanding the performance gains obtained from the addition of visual cues.

We study the performance of audio-only speech enhancement models as a function of what is being articulated, where we use \emph{visemes} as the basic unit of analysis.  A viseme consists of a cluster of phonemes that share the same place of articulation, and so visemes represent visually indistinguishable phonemes~\cite{neti2000audio}.  For example, the phonemes /uh/ and /w/ both map to a rounded vowel viseme, while phonemes /b/ and /m/ map to a viseme representing bilabial consonants.  We hypothesize that enhancement performance will vary depending on what is being said since certain sounds are more visually prominent than others.  Given the per-viseme audio-only enhancement performance, we then quantify the performance gains obtained from the addition of visual cues to the enhancement model.

We also hypothesize that the visual embeddings implicitly learned by the audiovisual model can be used for other visual speech tasks. We show that these visual embeddings can be used to discriminate visemes during continuous speech, e.g.,\ rounding lips, stretching lips, and visible teeth.  Our results show that audiovisual speech enhancement can be used as a self-supervision task for learning meaningful visual speech embeddings without relying on manual annotations.


\section{Audiovisual Enhancement Model}\label{sec_architecture}

\begin{figure}[t]
\centering
\def\svgwidth{\linewidth}
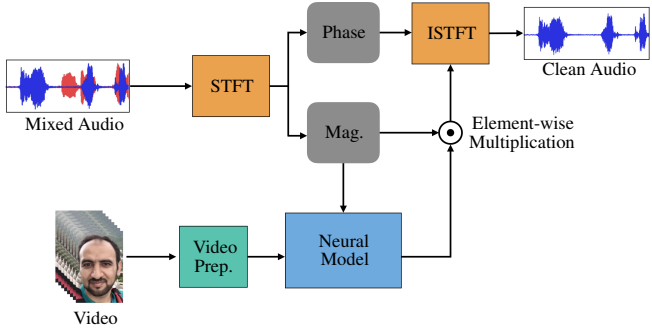
\caption{The audiovisual speech enhancement system.}\label{fig:model_system}
\end{figure}

Our architecture is shown in Figure~\ref{fig:model_system}. The neural enhancer receives two inputs: the squared magnitude of the short-time Fourier transform (STFT), i.e., the power spectrum, of the \emph{mixed} speech segment, and a video segment containing the corresponding pose-normalized gray-scale mouth images of dimension $w\times h \times t$.  To produce an \emph{enhanced} version of the input speech, the model predicts an ideal ratio mask (IRM), which we write as:

\begin{displaymath}
IRM(m,f)=\frac{|S{(m,f)}|^2}{|S{(m,f)}|^2 + |N{(m,f)}|^2}
\end{displaymath}
where $|S{(m,f)}|^2$ and $|N{(m,f)}|^2$ represent the power spectrums of the speech and noise signals at frame $m$ and frequency bin $f$.   Element-wise multiplying an IRM by the power spectrum of the mixed signal gives an optimal estimate, in the sense of the  minimum mean square error (MMSE), of the power spectrum of the clean signal~\cite{loizou2013speech}.

\subsection{Audiovisual Neural Model}
\begin{figure}[t]
\centering
\def\svgwidth{\linewidth}
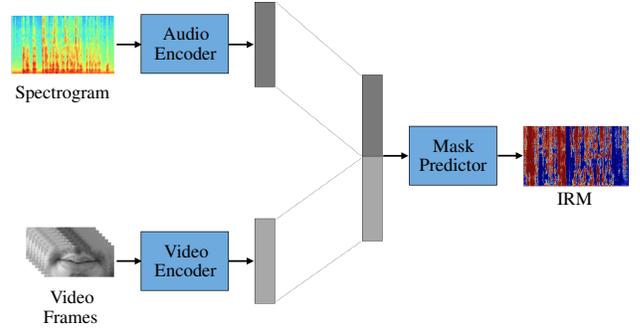
\caption{The audiovisual speech enhancement neural model.}\label{fig:model_architecture}
\end{figure}

The audiovisual neural model, shown in Figure~\ref{fig:model_architecture}, consists of three sub-networks: the audio encoder, the video encoder, and the mask predictor.

The audio encoder induces an embedding given the acoustic representation of the mixed input speech.  We experiment with both fully-connected- and LSTM-based audio encoders in this work. 
The video encoder induces an embedding given the video representation.  
The video encoder is based on the VGG-M architecture~\cite{chung2016out}, which consists of a series of conv-pool layers, followed by a series of fully-connected layers.
Finally, the mask predictor outputs an IRM given the concatenated multimodal embedding.  The mask predictor consists of a series of fully-connected layers followed by a fully-connected linear regression layer.

\section{Experimental Setup}\label{sec_data}

\subsection{Dataset}
We use an in-house audiovisual corpus containing around 68 hours (39,097 utterances) of speech from 600 gender-balanced speakers. The utterances are  queries for a digital assistant spoken in English with an American accent. 
The audio is sampled at 16kHz using a 16-bit PCM encoding.  The video has a frame rate of 60Hz and a resolution of $720 \times 1280$.  We randomly split the dataset into gender-stratified partitions using a 80/10/10 rule.   The resulting splits consist of 480 speakers (29,415 utterances, 52 hours) for training, 60 speakers (4,650 utterances, 8 hours) for validation, and 60 speakers (5,032 utterances, 8 hours) for testing.

\subsection{Details}
Mixed utterances for training are created on-the-fly by mixing the target utterance with a random utterance from a different speaker in the training set. The mixtures used for the validation and test sets are fixed and are created using speakers from their respective partitions to ensure that the model does not see any of the validation or test speakers during training. All of the samples are mixed at signal-to-noise-ratio (SNR) of 0dB.

\textbf{Features:} We train our models using 200ms audiovisual segments.   Audio features represent the squared magnitude of the STFT of the mixed input signal extracted using a 25ms Hamming window with a hop size of 10ms. Visual features represent a sequence of $128\times96$ cropped gray-scale images of the mouth region for the target speaker extracted using dlib~\cite{king2009dlib}.

\textbf{Training:} The neural networks are trained with the ADAM optimizer using a learning rate of 1\(e\)-4 for a maximum of 100 epochs.  We monitor the validation performance during training and apply early stopping when the validation loss converges. The loss functions that we use are described in Section~\ref{subsub_baseline}.

\textbf{Architecture:} The video encoder is based on the VGG-M architecture~\cite{chung2016out} and consists of five convolutional blocks followed by three linear blocks. Each convolutional block consists of a \(3\times3\) convolution layer, followed by batch normalization, a ReLU non-linearity, and \(2\times2\) max-pooling. We use 96 filters in the first convolution layer and 128 filters each of the proceeding convolution layers.  Each linear block consists of a linear layer, followed by a ReLU non-linearity.  We use width sizes of 1024, 512, and 256 for the first, second, and third linear layer, respectively.   The audio encoder consists of three 512-dimensional  fully-connected (or LSTMs) layers.  Finally, the mask predictor is made up of three linear blocks, each with a width of 512. The hyper-parameters for the audio encoder and the mask predictor are chosen based on validation performance.

\textbf{Metrics:} Three metrics are used to evaluate the performance of the audiovisual enhancers: mean absolute error (MAE), signal-to-noise-ratio (SNR), and perceptual evaluation of speech quality (PESQ). SNR is used for measuring background noise reduction, while PESQ is used for measuring perceptual speech quality~\cite{rix2001perceptual}. We use SNR and PESQ for measuring the performance of the enhanced reconstructed signals at the utterance level and use MAE for measuring the performance of the predicted IRMs at the 200ms segment level. We use the ground-truth clean reconstructed signals as reference when computing SNR and PESQ\@.

\section{Results and Analysis}\label{sec_results}

\subsection{Baseline}\label{subsub_baseline}

\begin{table}[t]
\caption{Enhancement performance obtained for each setup and loss function. \textbf{SNR:} signal-to-noise-ratio in decibel (dB), \textbf{PESQ:} perceptual evaluation of speech quality, \textbf{A:} Audio-only enhancer, \textbf{AV:} audiovisual enhancer, \textbf{FC:} fully connected,  \textbf{LSTM:} Long short-term memory}\label{tab:results_1}
\vspace{3mm}
 
  \begin{tabular}{cccccc}
    \toprule
    \multirow{2}{*}{ \begin{tabular}{@{}c@{}}\textbf{Audio} \\ \textbf{Encoder}\end{tabular}} & \multirow{2}{*}{ \textbf{Loss} } & \multicolumn{2}{c}{\textbf{SNR}} & \multicolumn{2}{c}{\textbf{PESQ}}\\
    & \textbf{} & \textbf{A} & \textbf{AV} & \textbf{A} & \textbf{AV}\\
    \midrule
    \multirow{3}{*}{FC}                 &MSE         & 4.21 & 7.53 & 2.70  & 2.88 \\
                                        &MAE         & 4.28 & 8.10 & 2.56  & 2.87 \\
                                        &MAE+Cosine  & 4.62 & 8.07 & 2.67  & 2.90 \\
    \midrule
    \multirow{3}{*}{LSTM}               &MSE         & 4.61 & 8.42 & 2.67 & 2.92 \\
                                        &MAE         & 4.63 & 8.56 & 2.58 & 2.90 \\
                                        &MAE+Cosine  & 5.17 & 8.87 & 2.73 & 2.95 \\
    \bottomrule
  \end{tabular}
\end{table}

We seek to build a strong baseline model to be used for further analyses. We fix the visual encoder and study how changes to the audio encoder and the loss function affect the audiovisual enhancement performance.  For the audio encoder, we compare the performance of the fully-connected (FC)-based encoders to that of LSTM-based encoders. We compare three regression-based loss functions: mean squared error (MSE), mean absolute error (MAE), and a hybrid loss function that combines MAE with the cosine distance. 

MSE is a common loss function used in regression problems. Minimizing the MSE is equivalent to maximizing the log-likelihood of data with a unimodal Gaussian distribution. Upon further inspection of the distribution of the training targets, i.e.,\ the IRMs, we find that it does not resemble a unimodal Gaussian. Instead, the distribution of our training targets is bimodal, with a very large peak at zero (sparse labels) and a second smaller peak at one. The MSE solution in this case, which is the conditional mean of the distribution, will be between the two peaks, shifted toward the higher peak, at zero. This results in predicting blurry masks, which is consistent with observations about using the MSE loss in computer vision applications~\cite{zhang2016colorful}.

Using the MAE loss function can mitigate some of the limitations incurred from using the MSE loss function by encouraging the prediction of sharper IRMs~\cite{afouras2018conversation,afouras2019my}.  One remaining limitation with using both MSE and MAE loss functions is the assumption that the individual components of the IRM vector are statistically independent.   To address this limitation, we propose using a joint loss function that combines MAE with the cosine loss function.   The cosine loss measures the distance between two \emph{entire vectors} instead of measuring the distance between individual vector components.  The cosine distance, however, cannot be used as a standalone loss, as it minimizes the angle between two vectors irrespective of their magnitudes. This can result in IRM vectors with magnitudes beyond the masks' boundaries, i.e.,\ zero and one.   Therefore, we use the following hybrid loss of the MAE and cosine distance to optimize the angle between the ground truth and inferred IRM vectors while bounding their magnitude values to be between zero and one:
$$\mathcal{L}_\text{hybrid} = \mathcal{L}_\text{MAE}+\alpha~\mathcal{L}_\text{cos}$$
where $\alpha$ is a trade-off parameter that we set to $0.5$ in our experiments.

Table~\ref{tab:results_1} gives a summary of the results obtained from our baseline experiment.  The results show that considerable gains are achieved using an audio-only enhancer (columns labeled \textbf{A}), which was not expected a priori. One reason for this is that although the target mixture for the noisy signal was 0dB SNR, mixtures of 1dB emerged due to short pauses in the target and background speech. This 1dB difference between target and background acoustic speech gives the network a clue for enhancing the target speaker, even without visual cues.  That said, the results show that the addition of visual cues still provides improvement in performance for all setups.  The results also show that using an LSTM-based audio encoder yields better performance compared to FC-based encoders. 
Finally, the results show that using the proposed hybrid loss function gives improvements over using MSE for a majority of the setups.

\subsection{Viseme-specific Relative Improvements}

\begin{table}[t]
    \caption{The phoneme-viseme mapping used in our work and the enhancement improvements gained per viseme due to the addition of visual cues. $\%\Delta$ notes the percentage decrease in MAE (higher is better).}\label{table:PHM}
    \vspace{3mm}
    \centering{
        \begin{tabular}{cccc}
                \toprule
                \textbf{Viseme cluster}              & \textbf{Viseme}                    &  \textbf{Phoneme}  &  \textbf{$\%\Delta$}       \\
                \midrule
                Lip rounded vowels                   &   \multirow{2}{*}{ /V1/ } & /aa/~/ah/~/ao/              & \multirow{2}{*}{26.6}\\
                level 1                              &                           & /aw/~/er/~/oy/ \\
                 \midrule
                Silence                              & /SIL/                     & /sil/~/sp/                  & 25.0\\
                \midrule
                Bilabial                             & /P/                       & /p/~/b/~/m/                 & 23.1\\
                 \midrule
                Lip stretched vowels                 & \multirow{2}{*}{/V3/}     & /ae/~/eh/~/ey/              & \multirow{2}{*}{20.0}\\
                level 1                              &                           & /ay~/y/\\
                 \midrule
                \multirow{2}{*}{Palato alveolar}     & \multirow{2}{*}{/SH/}     & /sh/~/zh/~/ch/              & \multirow{2}{*}{19.0}\\
                                                     &                           & /jh/\\
                \midrule
                Alveolar semivowels                  & /L/                       & /l/~/el/~/r/                & 17.6\\
                 \midrule
                Lip rounded vowels                   &  \multirow{2}{*}{/V2/ }   & /uw/~/uh/~/ow/              & \multirow{2}{*}{16.7}\\
                level 2                              &                           & /w/ \\
               \midrule
                \multirow{2}{*}{Velar}               & \multirow{2}{*}{/G/}      & /g/~/ng/~/k/                & \multirow{2}{*}{15.4}\\
                                                     &                           & /hh/\\
                \midrule
                Alveolar fricative                   & /Z/                       & /z/~/s/                     & 14.3\\
                \midrule
                \multirow{2}{*}{Alveolar}            &\multirow{2}{*}{ /T/ }     & /t/~/d/~/n/                 & \multirow{2}{*}{13.3}\\
                                                     &                           & /en/\\
                \midrule
                Dental                               & /TH/                      & /th/~/dh/                   & 13.3\\
                \midrule
                Labio-Dental                         & /F/                       & /f/~/v/                     & 13.3\\
                 \midrule
                Lip stretched vowels                 &  \multirow{2}{*}{/V4/}    & \multirow{2}{*}{/ih/~/iy/}  & \multirow{2}{*}{12.5}\\
                level 2                              &                           &    \\ 
            \bottomrule
        \end{tabular}
    }
\end{table} 

In this section, we investigate whether the visual features improve the speech enhancement model by simply providing it with voice activity features, i.e.,\ speech/silence, or by providing the model with more fine-grained information about what is being articulated. We compare the per-viseme improvements of the audio-only and audiovisual speech enhancement models in terms of the MAE between the inferred and ground truth IRMs. The per-viseme performance is obtained using three steps. First, we apply an in-house ASR model to all test utterances to estimate frame-phoneme alignments. Next, we cluster the phonemes into visemes following the phoneme-viseme mapping in Table~\ref{table:PHM}, which is a modified version of the mapping used in~\cite{neti2000audio}. Finally, we compute the MAE between the predicted mask and the IRM in the test set and report the per-viseme performance.

The results in Table~\ref{table:PHM} show that the addition of visual cues results in performance improvements for all visemes.
However, these performance gains vary based on what is being articulated. 
For instance, we see improvements for the viseme /SH/, which sounds like noise acoustically but is easy to classify visually. We also see different gains for the phoneme /m/, which is mapped to the viseme cluster /P/, and phoneme /n/, which is mapped to the viseme cluster /T/. Both of these phonemes sound similar acoustically but look different visually. 


\subsection{Viseme Classification}

\begin{table}[t]
\caption{Viseme classification performance obtained using visual embeddings extracted from the video encoder in the enhancement model.}\label{tab:results_3}
  \vspace{3mm}
  \centering
  \begin{tabular}{cc|cc}
    \toprule
    \textbf{Viseme}    &\textbf{Recall (\%)} & \textbf{Viseme}    &\textbf{Recall (\%)} \\
    \midrule
      /SIL/ & 84.3  &  /V4/  & 24.0 \\
      /SH/  & 68.7  &  /L/   & 20.6\\
      /P/   & 65.1  &  /TH/  & 19.0\\
      /F/   & 50.0  &  /G/   & 9.5\\
      /Z/   & 43.1  & /T/   & 4.2   \\
      /V1/  & 39.6  & Average & \textbf{33.5}\\
      /V3/  & 31.1  & Chance             & 7.7\\
      /V2/  & 28.5  & & \\
    \bottomrule
  \end{tabular}
\end{table}

In this section, we investigate if audiovisual speech enhancement can be used as a self-supervised task for learning meaningful visual embeddings that can be used in other visual speech applications. Given the trained audiovisual speech enhancement model from our previous experiment, we disconnect the video encoder and use it as a general feature extractor. We use these extracted features to train a logistic regression model for viseme classification. For training the logistic regression model, we further split the test set used for evaluating the audiovisual speech enhancement model into training, validation, and test sets following a speaker-independent 80/10/10 split rule. This approach ensures two things: (1) the speakers used for training the enhancement models are different from those used in our analysis; and (2) the logistic regression model is trained, validated, and tested on speaker independent partitions. The $C$ hyper-parameter of the logistic regression model is tuned using the validation set. The performance is evaluated in terms of recall per viseme.

Table~\ref{tab:results_3} shows the viseme classification performance obtained when using the visual embeddings as features for a simple logistic regression viseme classifier. We find that the visual embeddings are discriminative toward visemes, giving an overall unweighted average recall of 33.5\%, where 7.7\% is the chance performance. We find that our classifier predicts apparent visemes, such as /Z/, /F/, /P/, and /SH/, relatively accurately compared to predicting visemes articulated more towards the back of the mouth, such as /T/ and /G/. The trends that we observe for viseme prediction performance using visual embeddings are similar to those observed in viseme classification tasks. As a benchmark, we obtained an unweighted accuracy of 49.2\% using a separate VGG-M neural network trained from scratch specifically to detect visemes, which suggests that our self-supervised visual features are able to close a large proportion of the performance gap. This demonstrates the efficacy of audiovisual speech enhancement as a self-supervised task for learning strong visual features.

\section{Conclusion}\label{sec_conclusion}
In this paper, we shed light on how an audiovisual speech enhancement model utilizes visual cues to improve the quality and intelligibility of a target speech signal. We showed that the performance of enhancement models varies depending on what is being articulated. We also showed that the addition of visual cues provides non-consistent gains in performance depending on what is being articulated. Further, we demonstrated the effectiveness of audiovisual speech enhancement as a self-supervision task for learning meaningful visual embeddings for visual speech applications.

\bibliographystyle{IEEEbib}
\bibliography{mybib}

\end{document}